\documentclass{article}
\usepackage[utf8]{inputenc}
\usepackage[margin=2cm]{geometry}
\usepackage{fullpage,enumitem,amssymb,amsmath,amsthm,xcolor,cancel,gensymb,hyperref,graphicx}

\usepackage{indentfirst}
\usepackage{algorithmic}
\usepackage[ruled,vlined]{algorithm2e}
\usepackage[caption=false]{subfig}

\usepackage{natbib}
\usepackage{bbm}
\newtheorem{thm}{Theorem}
\newtheorem{lem}[thm]{Lemma}

\newtheorem{defn}{Definition}

\DeclareMathOperator*{\argmax}{arg\,max}
\DeclareMathOperator*{\argmin}{arg\,min}

\DeclareMathOperator*{\bbE}{\mathbb{E}}
\DeclareMathOperator*{\bbR}{\mathbb{R}}

\DeclareMathOperator*{\bbP}{\mathbb{P}}

\DeclareMathOperator*{\util}{\textit{U}}

\DeclareMathOperator*{\hp}{\widehat{\textit{p}}}
\DeclareMathOperator*{\hq}{\widehat{\textit{q}}}
\DeclareMathOperator*{\htheta}{\widehat{\theta}}
\DeclareMathOperator*{\halpha}{\widehat{\alpha}}
\DeclareMathOperator*{\hbeta}{\widehat{\beta}}

\DeclareMathOperator*{\hO}{\widehat{\textit{O}}}

\DeclareMathOperator*{\heta}{\widehat{\eta}}

\DeclareMathOperator*{\calX}{\mathcal{X}}

\DeclareMathOperator*{\calS}{\mathcal{S}}

\title{Addressing Distribution Shift in RTB Markets\\ via Exponential Tilting}

\author{ Minji Kim ~~~~~~~  Seong Jin Lee \\ 
\texttt{mkim5@unc.edu} ~~~ \texttt{slee7@unc.edu} \\
University of North Carolina, Chapel Hill \and
Bumsik Kim \\
\texttt{bumsik@moloco.com} \\
    Moloco Inc.
}

\date{ }

\begin{document}

\maketitle
\vspace{-0.5cm}

\begin{abstract}

In machine learning applications, distribution shifts between training and target environments can lead to significant drops in model performance. This study investigates the impact of such shifts on binary classification models within the Real-Time Bidding (RTB) market context, where selection bias contributes to these shifts. To address this challenge, we apply the Exponential Tilt Reweighting Alignment (ExTRA) algorithm, proposed by \cite{maity2023understanding}.
This algorithm estimates importance weights for the empirical risk by considering both covariate and label distributions, without requiring target label information, by assuming a specific weight structure. The goal of this study is to estimate weights that correct for the distribution shifts in RTB model and to evaluate the efficiency of the proposed model using simulated real-world data.


\end{abstract}

\section{ Introduction }
\label{sec:intro}
Under the dynamic nature of machine learning models, distribution shifts are a common issue where the source and target data distributions differ significantly, especially when deployed in new environments or making predictions in online settings.
This discrepancy can lead to a significant drop in performance when a model trained solely on source data is applied to target data. To illustrate, consider labeled data $(X,Y)$ with covariate $X$ and label $Y$ pairs. Distribution shifts can take various forms; for example, according to \cite{zhang2023dive}, Section 4.7, these include covariate shifts (changes in the distribution of $X$), label shifts (changes in the distribution of $Y$), and shifts where both covariates and labels deviate, such as concept shifts.

To address distribution shifts, weighting methods introduce importance weights into the empirical risk (\cite{shift}).
Assume that the joint data are drawn from the source distribution $P$, and a model $f$ is designed to minimize the empirical risk, defined as
\begin{align}\label{eq:risk}
    {R}_{P}(f) = {\bbE}_{ P}l(f(X),Y) \approx \frac{1}{N}\sum_{i=1}^N l(f(X_i), Y_i),
\end{align}
where $l$ denotes the loss function. 
Assume further that the target data come from a different distribution $Q$ and there exists probability density functions (p.d.f.s) $p$ and $q$ corresponding to the measures $P$ and $Q$, respectively. Note that the expected risk in the target domain can be written as:
\begin{equation}
    \begin{aligned}
        \label{eq:intro-risk-is}
    {R}_Q(f) &= {\bbE}_{Q } l(f(X),Y) = \int l(f(X),Y) dQ\\
    & = \int l(f(X),Y) \frac{q(X,Y)}{p(X,Y)}dP = {\bbE}_{P } \Big [ l(f(X),Y)\frac{q(X,Y)}{p(X,Y)} \Big ].
    \end{aligned}
\end{equation}
To ensure the model performs effectively on data drawn from the target distribution, performance can be improved by minimizing the weighted empirical risk:
\begin{align}
\label{eq:intro-loss-is}
    \frac{1}{N} \sum_{(X_i,Y_i)\sim P} l(f(X_i),Y_i) \frac{q(X_i,Y_i)}{p(X_i,Y_i)}.
\end{align}
Therefore, learning the importance weight ${q(X,Y)}/{p(X,Y)}$ can effectively adjust for the shift between the source and target distributions. 
\begin{figure}
\centering
\includegraphics[width=.85\textwidth]{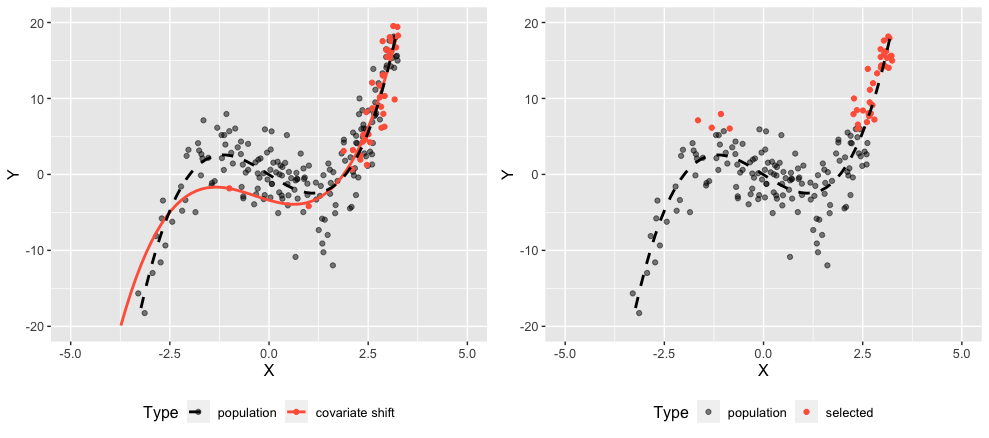} 
\caption{Illustration of distribution shift: (Left) covariate shift and (Right) sample selection bias.}
\label{fig:cov}
\end{figure}

Estimating these weights in practice is a challenging task that depends on the specific nature of the distribution shift.
Figure~\ref{fig:cov} illustrates two distinct scenarios of distribution shift. 
In the left panel, the source dataset used for training the model is highlighted in orange to demonstrate an example of covariate shift, where the relationship between $Y$ and $X$ remains consistent between the source dataset and the target dataset. However, in the target dataset, the range of $X$ is broader, shifting significantly toward smaller values.
Under covariate shift, weights are given as $q(X)/p(X)$ by using the marginal distribution of $X$ without observing $Y$.
When the goal is to model the population (which is our target set), more weight will be assigned to the loss associated with smaller values of $X$.
On the other hand, selection bias presents another interesting scenario, where sample selection is correlated with $Y$, leading to changes in both covariate and label distributions. For instance, the right panel illustrates a case where sample selection for the target dataset occurs primarily when the values of $Y$ are large.
This distorts the conditional distribution of $Y|X$, thereby significantly affecting the model.

In many real-world prediction problems, target label information is often limited or unavailable during training, making the covariate shift assumption particularly relevant. Various studies have explored weighting methods under the covariate shift. For example, \cite{JMLR:v8:sugiyama07a} examines weighted empirical risk minimization, while \cite{SHIMODAIRA2000227} discusses maximum weighted likelihood estimation under a parametric model.
Additionally, \cite{conformal} introduces a method for handling uncertainty in predictions when covariate shift is present.
The problem of interest in our study, which will be further detailed in Section~\ref{sec:rtb}, focuses on binary classification within the Real-Time Bidding (RTB) market. This problem involves challenges such as class imbalance and selection bias. As illustrated in Figure~\ref{fig:cov}, the distribution shift caused by selection bias cannot be adequately addressed by covariate shift assumptions, necessitating the estimation of weights that consider both covariates and labels. 
One such approach is the Exponential Tilt Reweighting Alignment (ExTRA) algorithm introduced by \cite{maity2023understanding}, which is particularly attractive because it estimates weights by considering both covariates and labels without requiring target label information. However, this method involves a trade-off, as it relies on a specific weight structure between the source and target domains, motivated by the exponential tilt model.
The goal of our study is to examine the specific characteristics of binary classification in the RTB market and to apply ExTRA algorithm to deal with distribution shift in RTB market model.


The rest of the paper is organized as follows: Section~\ref{sec:rtb} introduces the binary classification problem in the RTB market. Section~\ref{sec:tilt} discusses the exponential tilt model and the estimation algorithm proposed by \cite{maity2023understanding}. In Section~\ref{sec:real}, we evaluate the weighting approach using simulation settings that reflect real market data. Section~\ref{sec:con} concludes and suggests future directions.

\section{Real-Time Bidding market model} 
\label{sec:rtb}
RTB is a digital advertising method where advertisement impressions are auctioned in real-time, often in milliseconds, as a user loads an ad slot.
What makes this mechanism intriguing from a machine learning perspective lies in predicting, in split seconds, which advertisement will resonate most with the profile of the user based on various features. 
As the RTB market dynamics fluctuate and user behaviors shift over time, the underlying data distributions can change, introducing challenges for predictive models that were trained on previous data.
The RTB market, therefore, provides a rich context to explore the intricacies of managing distribution shifts.
In this study, we consider market models for predicting a binary outcome indicating the utility of a bid opportunity, such as app installations, denoted as $U \in \{0, 1\}$.
The features associated with each observation are denoted as covariate $X$. Note that while we referred to the target random variable as $Y$ in Section~\ref{sec:intro}, from this section onward, we use $U$ to emphasize the binary target variable.

Two significant characteristics in this dataset are class imbalance and selection bias. Class imbalance occurs because successful outcomes ($U=1$) are much rarer compared to unsuccessful ones ($U=0$). 
For instance, expected app install probabilities often fall in the range of $10^{-3}$ to $10^{-5}$ or could be even lower. 
Note that the loss function used in empirical risk minimization, as defined in \eqref{eq:risk}, treats misclassification errors from both classes with equal importance. This leads classifiers to be biased towards the majority class.
For example, \cite{imbalance_bias} points out that skewed class distribution in source data tends to produce unreliable class probability estimates, especially for minority class instances.
In the context of estimation problems using Monte Carlo samples, importance sampling is widely used to improve the efficiency of rare event estimation, such as in reliability engineering for estimating failure probabilities (\cite{BIONDINI201529}, \cite{CHIRON2023109238}).
In the machine learning context, particularly for predictive models, a more delicate approach involves giving greater weight to the loss associated with rare events to systematically reshape the optimization objective.
For instance, \cite{lin2017focal} introduced Focal Loss, which adjusts the loss function to focus more on difficult-to-classify examples in dense object detection, while \cite{calibration} presented a method for calibrating probability with undersampling. Additionally, \cite{li2021tilted} introduced (exponentially) tilted empirical risk minimization criteria that allow for emphasizing misclassifications.

\begin{figure}
\centering
\includegraphics[width=.6\textwidth]{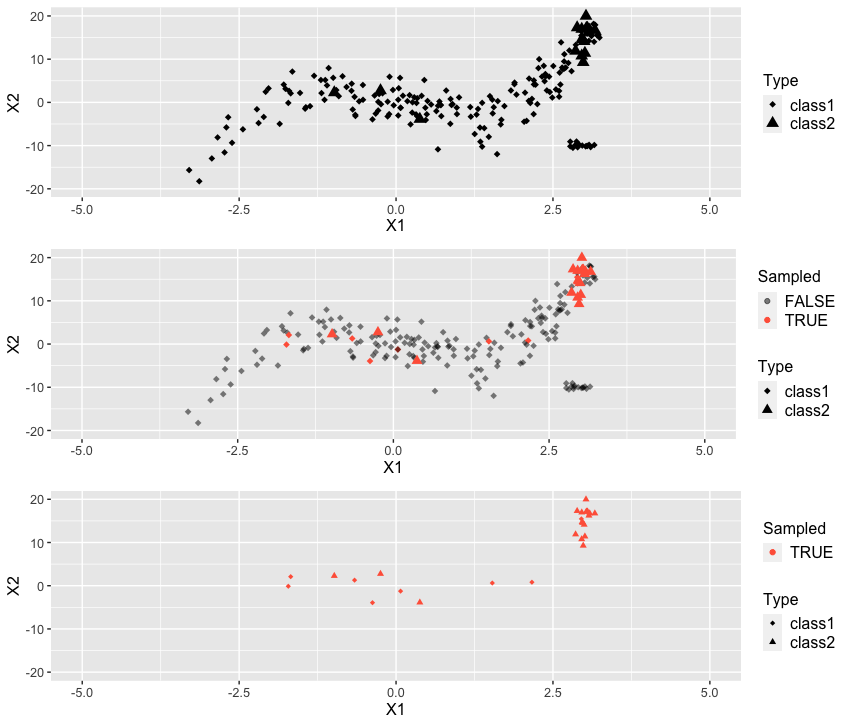} 
\caption{Illustration of selection bias in binary classification}
\label{fig:sel}
\end{figure}

Another characteristic of this dataset is sample selection bias, which occurs when the sample selection process depends on the target variable $U$. 
To illustrate, consider data $(X,U)$, with a binary sample selection variable $V$; $V=1$ indicates a selected sample (i.e., the sample is observed and consists of the source data) and $V=0$ otherwise. 
Assuming that the true joint distribution of $X$ and $U$ follows $p_o$, the source p.d.f., $p$, is written as
\begin{align*}
    p(X,U) = p_o(X, U | V=1) = p_o( U | X, V=1) p_o(X | V=1),
\end{align*}
while the target p.d.f., $q$, is the true landscape
\begin{align*}
    q(X,U) = p_o(X,U) = p_o( U | X) p_o(X). 
\end{align*}
Note that if $V$ and $U$ are independent conditionally on $X$, then the scenario could be classified as a covariate shift, as the distribution shift only depends on that of the covariate $X$. However, when $V$ depends on $U$, the selection bias introduces a more complex form of distribution shift.
Figure~\ref{fig:sel} provides a visual illustration of selection bias in a binary classification problem, compared to what we have observed from the continuous $Y$ case in Figure~\ref{fig:cov}. The top panel represents the entire population, where we are interested in distinguishing between two types, $U=0$ (class1; circles) and $U=1$ (class2; triangles). Suppose that we only observe a subset of samples, depicted in orange points in the bottom panel. If the selection of these samples (i.e., whether to observe the sample) is not independent of the distribution of $U$, the model's prediction of $U$ solely based on the observed samples, $P(U|X, V=1)$, could be significantly differ from the true prediction $P(U|X)$. 
In this simplistic example, predicting underlying distribution of $U$ based solely on the observed sample would be nearly impossible for any model. Nevertheless, if the goal of our model is to estimate population quantities (e.g., population mean) based on the observations, we must consider appropriate strategies, such as exploration-exploitation algorithms (\cite{bandit}) or using weights to match the distributions.

To describe the presence of selection bias in RTB market model, we define $U_i\in \{0,1\}$ the \textit{utility} of request $i$ and $M_i \in \mathbb{M}$ the \textit{market condition} of request $i$ as the random variables with proper distribution function and some measurable space $(\mathbb{M}, \mathcal{M})$. The estimated utility given a specific user context for the request $\bbE (\util_i | X_i)$ can be subsequently utilized to determine the optimal bid price in an optimization process. 
When $M_i$ denotes the winning auction price, 
$\util_i$ is observable only when our bidding price $b$ exceeds $M_i$, meaning that we win the auction. Thus, a sample is chosen when $M_i<b$. An estimator derived from the source set is given by
$p_i := \bbE_{\text{win}}(\util_i | X_i) = P(\util_i =1| X_i, M_i<b)$. 
If we denote $\pi_i := \bbE_{\text{total}} (\util_i | X_i ) $ to be the true probability in the target domain, the following relationship holds,
\begin{align}\label{eq:conditional_p}
    p_i &= {\bbP}_{\text{win}}(U_i=1 | X_i)= \frac{\pi_i \cdot F_{i,1}(b)}{ F_i(b)},
\end{align}
where $F_{i,1}, F_{i,0}$ are conditional c.d.f.s of $M_i$ given $X_i$ and $U_i=1$ or $U_i=0$, respectively, and $F_i(b) =(1-\pi_i) F_{i,0}(b) + \pi_i F_{i,1}(b)$ is the conditional c.d.f.\ of $M_i$ given $X_i$. 
As \eqref{eq:conditional_p} indicates, $p_i$ converges to $\pi_i$ as $b \to \infty$. This means that there will be no bias if we win all bids. For any bid price $b$, $p_i$ equals $\pi_i$ when $F_{i,1}(b) = F_{i,0}(b)$. 
In conclusion, the dependence between market condition and utility can introduce selection bias, further increasing the gap between the observed distribution and the target population distribution.
This underscores the need for appropriate handling of such shifts, not only when predicting population distributions but also when assessing model performance in dynamic environments, such as changes in the total bidding budget or other real-time fluctuations.


\section{The Exponential Tilt Model}
\label{sec:tilt}

In this section, we examine an approach that utilizes importance weights for empirical risk, aimed at classifying utility in the target domain that differ from the source domain used for training. The model seeks to enhance prediction performance in target domains by assigning greater weight to relevant samples, rather than applying uniform source weights.
This section of our study closely follows the methods outlined in \cite{maity2023understanding}, and we provide detailed explanations from that study to ensure our discussion is self-contained.

\textbf{Problem setup} 
We consider a binary classification problem where $U$ takes values of $0$ or $1$. This setup can be easily extended to a $K$-class classification problem. Let $\mathcal{X}$ and $\mathcal{U}$ denote the space of features and the set of possible utility values, respectively. We assume the existence of probability measures and associated densities for both the source and target datasets. Specifically, let $P$ and $Q$ represent the probability distributions on $\mathcal{X} \times \mathcal{U}$, with $p$ and $q$ being the corresponding densities for the source and target domains.
The data we train on consists exclusively of `winning' bids. Consequently, our source domain is represented by a labeled dataset $\{X_{\{W,i\}}, U_{\{W,i\}}\}_{i=1}^{n_W}$, corresponding to these winning bids, where the subscript `W' denotes winning bids. In contrast, the target domain includes all bid opportunities—both winning and losing. In this domain, only unlabeled samples $\{X_{\{T,i\}}\}_{i=1}^{n_T}$ are available, with the subscript `T' standing for the total set of bids. The objective of the ExTRA algorithm in \cite{maity2023understanding} is to learn weights for samples from the source domain, enabling the weighted source distribution to mimic the target distribution. To make the learning of the weight function feasible, we introduce the following additional constraints on the domains.

\textbf{The exponential tilt model}  We assume that there is a vector of sufficient statistics $T: \calX \rightarrow \bbR^p$ and the parameters $\{\theta_i\in \bbR, \alpha_i\in \bbR\}_{i\in \{0,1\}}$ such that 
\begin{align} \label{eq:tilt}
    \log \frac{q(x,U=i)}{p(x,U=i)} = \theta_i^\top T(x) + \alpha_i, \  i\in \{0,1\}.
\end{align}
This assumption implies that $q(x,U=i)$ belongs to the exponential family with base measure $p(x,U=i)$ and sufficient statistics $T$ when normalized. We refer to \eqref{eq:tilt} as the exponential tilt model. It suggests the weights characterizing the distribution shift between the source and target datasets are given by
\begin{align}
\label{eq:weight}
    w(x,i) = \exp(\theta_i^\top T(x) + \alpha_i), \ i\in \{0,1\}.
\end{align}
The marginal density of the features in the target domain, $q_X$, is then represented as
\begin{align}\label{eq:q_iden}
  q_{X}\{x\} = \sum_{i\in\{0,1\}}p(x,U=i)\exp{(\theta_i^\top T(x) + \alpha_i )}.
\end{align}

The choice of \( T \) is crucial and requires careful consideration of several factors. The rationale behind selecting \( T \) is that it operates jointly with \( U \) to partition the dataset into subgroups. $T(X)$ is defined such that for each fixed \( t \),
\begin{align}\label{eq:T}
    (X_W,Y_W) | \{ T(X_W)=t, U_W = i \} \approx_d (X_T,Y_T) | \{ T(X_T)=t, U_T = i \}
\end{align}
holds.
Within each subgroup, the distribution shift described in \eqref{eq:weight} should enable effective matching from the source to the target domain. To achieve this, $T(X)$ jointly with $U$ should create partitions that are detailed enough to capture the necessary variation between the source and target domains, while ensuring that the subgroups meet certain conditions to maintain the identifiability of the parameters \( (\theta_i, \alpha_i)_{i\in\{0,1\}} \). For example, Lemma~\ref{lem:iden} is introduced to support the identifiability condition.

\begin{defn}[anchor set]
    A set ${\calS}_i \subset \calX$ is an anchor set for class $i$ if $p\{x,U=i\} > 0$ and $p\{x,U=j\}=0\  \forall j \neq i $
\end{defn}
\begin{lem}[Proposition 4.2 of \cite{maity2023understanding}]\label{lem:iden}
    If there are anchor sets ${\calS}_i$ for all $K$ classes in the source domain, and $T(\calS_i)$ is $p$-dimensional, then there is at most one set of $\theta_i$'s and $\alpha_i$'s that satisfies eq~\eqref{eq:q_iden}.
\end{lem}

\textbf{Fitting the exponential tilt model}
With the formulation in \eqref{eq:T}, the optimization problem for distribution matching, constrained by the requirement that the marginal density integrates to one, is described by:
\begin{equation}\label{eq:optim}
\begin{aligned}
  \{{\htheta}_i, {\halpha}_i\}_{i\in\{0,1\}} \in &\argmin_{(\theta_i,\alpha_i)} \sum_{i\in\{0,1\}} D\big({\hq}_X\{x\}\| \hp\{x, U=i\} \exp{(\theta_i^\top T(x) + \alpha_i )} \big)\\
    &\text{subject to }  \int_{\mathcal{X}} {\sum_{i\in\{0,1\}} \hp\{x,U=i\}\exp(\theta_i^\top T(x)+\alpha_i) dx} =1
\end{aligned}
\end{equation}
The choice of $D$ as a metric to measure the distance between distributions is flexible. \cite{maity2023understanding} demonstrated that using the Kullback-Leibler (KL) divergence allows the optimization to proceed without the need to estimate $p(x, U=i)$ or $q(x)$. Instead, it requires a probabilistic classifier $\heta_W = \{\heta_{W,i}\}_{i\in \{0,1\}}$ trained on the source domain, which can be expressed as
\begin{align*}
    {\heta}_{W,i}(X) = \hp(U=i|X) = \hp(X,U=i)/\hp(X).
\end{align*}
The following Lemma~\ref{lem:optim} and Algorithm~\ref{alg:extra} are used to fit the exponential tilt model.
\begin{lem}{(Lemma 3.1 of \cite{maity2023understanding})}\label{lem:optim}
    If $D$ is the Kullback-Leibler (KL) divergence then optima in \eqref{eq:optim} is achieved at $\{(\htheta_i,\halpha_i)\}_{i\in\{0,1\}}$, where
    \begin{align*}
      \{({\htheta}_i,{\halpha}^*_i)\}_{i\in\{0,1\}}  \in \underset{\theta_i, \alpha^*_i}{\argmax}\ {\bbE}_{\widehat{Q}}\big[ \log \big\{  \sum_{i\in\{0,1\}} {\heta}_{W,i}(X) \exp(\theta_i T(X) + \alpha^*_i) \big\} \big]
      - \log \big\{  {\bbE}_{\widehat{P}}[ \exp(\theta_U T(X) + \alpha^*_U) ]\big\}
    \end{align*}
 and
    $\halpha_i = {\halpha}^*_i - \log \big\{  {\bbE}_{\widehat{P}}[ \exp(\htheta_U T(X) + \halpha^*_U) ]\big\}$.
\end{lem}

\begin{algorithm} \label{alg:extra}
\KwData{ \renewcommand\labelitemi{\small$\bullet$}
\begin{itemize}
\setlength\itemsep{.2em}
    \item Dataset: labeled source data $\{X_{\{W,j\}}, U_{\{W,j\}}\}_{j=1}^{n_W}$ and unlabeled target data $\{X_{\{T,j\}}\}_{j=1}^{n_T}$.
    \item Hyperparameters: learning rate $\eta>0$, batch size $B\in \mathbb{N}$, normalization regularizer $\lambda>0$.
    \item Probabilistic source classifier: $\heta_W: \calX \rightarrow \Delta^I$
\end{itemize} }
Initailize $\htheta_0$ at some value.
\\
\Repeat{converges}
{
Sample minibatchs $(X_{W,1}, U_{W,1}), \dots, (X_{W,B},U_{W,B}) \sim \widehat{P}$, and $(X_{T,1}, \dots, X_{T,B}) \sim \widehat{Q}_X$\\
Compute loss $\widehat{L}_t = \frac{1}{B}\sum_{j=1}^B\log\big\{ \sum_{i\in\{0,1\}} \heta_{W,i}(X_{T,i})\exp(\htheta_{i,t}T(X_{T,j}) + \widehat{\beta}_i)\big\}$ and normalizer $\widehat{N}_t = \frac{1}{B} \sum_{j=1}^B \exp(\htheta_{U_{W,j},t}T(X_{W,j}) + \widehat{\beta}_{U_{W,j},t})$\\
Objective $\widehat{O}_t = - \widehat{L}_t +\log(\widehat{N}_t) + \lambda \widehat{N}_t +\lambda \widehat{N}_t^{-1}$\\
Update $\htheta_{i,t+1}\leftarrow -\eta \partial_{\theta_i}\hO_t\{(\htheta_{i,t}, \hbeta_{i,t}), i= 0,1 \}$ and $\hbeta_{i,t+1}\leftarrow \hbeta_{i,t} - \eta \partial_{\beta_i} \hO_{t}\{(\htheta_{i,t}, \hbeta_{i,t}), i= 0,1 \}$
}

Estimated value $\{(\htheta_i, \widehat{\beta}_i)\}_{i\in\{0,1\}}$\\
$\halpha_i \leftarrow \widehat{\beta}_i - \log \widehat{N}(\{(\htheta_i, \widehat{\beta}_i)\}_{i\in\{0,1\}})$\\
\Return parameters $\{(\htheta_i, \halpha_i)\}_{i\in\{0,1\}}$ and the weight function $w(x,u) = \exp(\htheta_y T(x))+\halpha_y$
 \caption{Exponential Tilt Reweighting Alignment (ExTRA, Algorithm 1 of \cite{maity2023understanding}) }
\end{algorithm}

\textbf{ Model evaluation in the target domain } One may estimate the model performance in the target domain by reweighing the empirical risk in the source domain, as shown in eq~\eqref{eq:intro-risk-is}:
\begin{align*}
    {\bbE}_Q[l(f(X_T), U_T)] \approx \frac{1}{n_W} \sum_{j=1}^{n_W} l(f(X_{W,j},U_{W,j}))\widehat{w}(X_{W,j},U_{W,j}).
\end{align*}
This allows us to evaluate models in the target domain without labeled samples. 
Alternatively, estimated weights can also be utilized to fine-tune models for the target domain by minimizing the reweighted empirical risk,
\begin{align*}
    \widehat{f}_T \in \argmin_{f\in \mathcal{F}}\frac{1}{n_W} \sum_{j=1}^{n_W} l(f(X_{W,j},U_{W,j}))\widehat{w}(X_{W,j},U_{W,j}), 
\end{align*}
where $\mathcal{F}$ is a function space.

\section{Simulated Real Data Analysis}
\label{sec:real}

\begin{figure}[h!]
\centering
\subfloat[ Distribution of source (left) and target (right) datasets]{{\includegraphics[width=.75\textwidth]{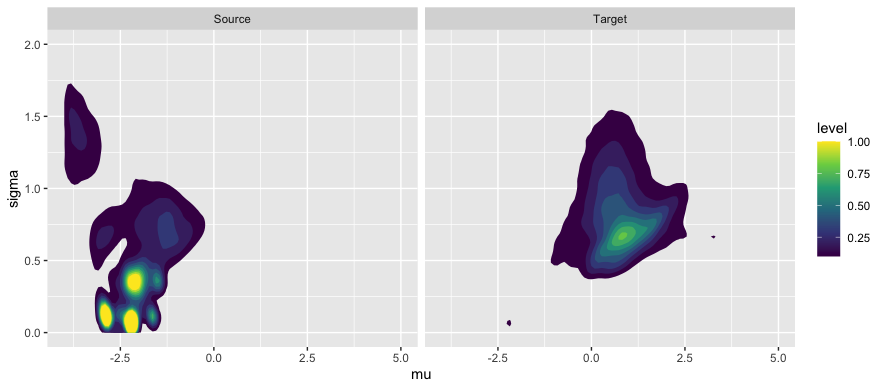} }}\hfill
\subfloat[ Distribution of source dataset conditioned on $U=0$ (left) and $U=1$ (right)]{{\includegraphics[width=.75\textwidth]{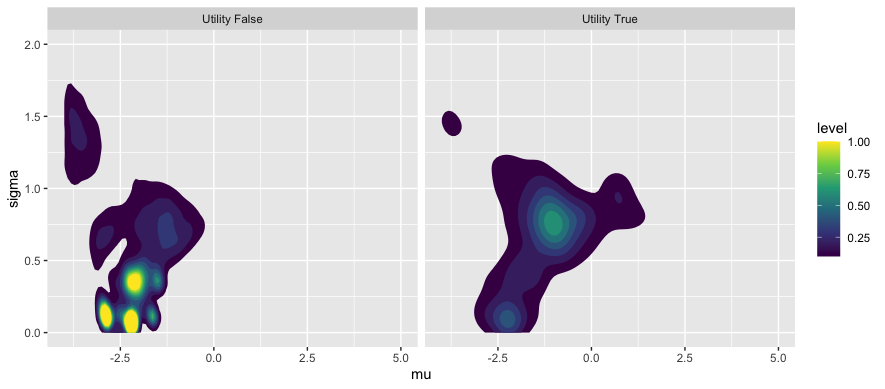} }}\hfill
\subfloat[Fitted weights for the source dataset conditioned on $U=0$ (left) and $U=1$ (right)]{{\includegraphics[width=.75\textwidth]{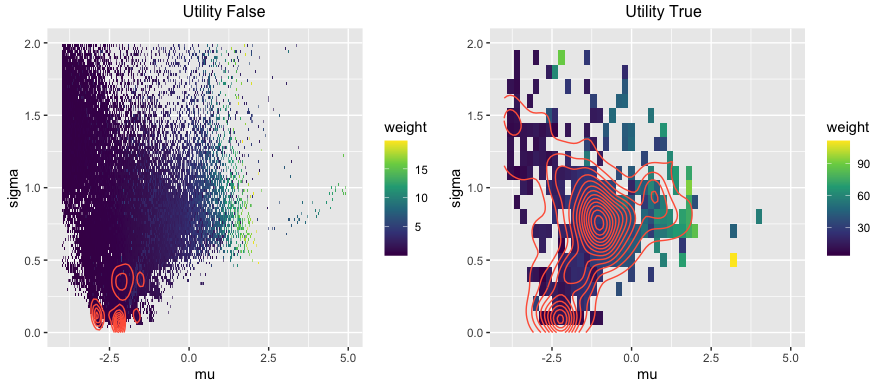} }}%
\caption{Distribution shifts and fitted weights in the RTB dataset:
(Top) Distribution of the source and target datasets, (Middle) Distribution of the source dataset conditioned on utility, and (Bottom) Distribution of the fitted weights for the source dataset.}
\label{fig:simul}
\end{figure}

In this section, we proceed with model fitting using Algorithm~\ref{alg:extra}, applied to data that closely resembles actual RTB bid records. The selection bias inherent in RTB market data occurs because the utility $U$ is only observable when an auction is won. This constraint limits our source dataset to instances where bids were successful, preventing us from training the utility prediction model on the full range of bid opportunities and features. As a result, our training is restricted to the subset of winning bids, while our objective is to estimate the expected utility $\bbE (U | X)$ across all bid opportunities. To achieve this, we aim to estimate the weights that can adjust the classification model trained on the winning bid set to accurately estimate the distribution of utilities for the total set of bids.

To proceed with the modeling, we first train a neural network model to estimate $p:=\bbP(U|X,\text{win})$ given a feature set $X$.
In the algorithm, $\heta_W$ corresponds to $\heta_{W,0}(X) = 1- p$ and $\ \heta_{W,1}
(X) = p$.
Next, we select the sufficient statistics $T(X)$ to be used, specifically choosing 20 variables derived from user and market features, including the estimated mean($\widehat{\mu}(X)$) and variance($\widehat{\sigma}(X)$) of the market condition.
To satisfy the conditions outlined in Lemma~\ref{lem:iden}, if a categorical variable is included in $T(X)$, there must be ample samples to ensure the presence of anchor set elements for each category.

Figure \ref{fig:simul} illustrates the distribution shift and the estimated weights defined in \eqref{eq:weight} within the simulated RTB dataset. 
In subplots (a) and (b), the colors represent levels of joint density, while in subplot (c), the colors correspond to the fitted weight values. 
In Figure \ref{fig:simul} (a), the joint distributions of features $m(X)$ and $\sigma(X)$ for both the source and target datasets are visualized, showing that the target distribution of $m(X)$ skews towards larger values compared to the source distribution.
Figure \ref{fig:simul} (b) displays the source dataset distributions conditioned on utility values $U=0$ and $U=1$, with the distribution for $U=1$ exhibiting a larger mean.
Figure \ref{fig:simul} (c) shows the fitted weights obtained using the ExTRA algorithm, with red contours indicating the joint density. These weights are designed to match the source dataset to the target dataset by accounting for the distribution shift.
In this specific case, the fitted weights assigned to the $U=1$ set, which represents the rare event, are larger, helping to address class imbalance by giving more weight to the rare set.
The distribution of the weights shows a clear preference for regions with larger means, which aligns with the observed distribution shifts and illustrates how the estimated weights from this approach performs.

\section{Conclusion}
\label{sec:con}
In this study, we addressed the challenge of binary classification in the RTB market, where selection bias induces significant distribution shifts. To mitigate these shifts and improve model performance in new target environments, we employed the exponential tilt model to estimate the weights that account for distribution shifts. This approach could lead to a more effective alignment between the source and target distributions.
However, further validation through studies on the selection and justification of $T(X)$ and testing the model across broader and more complex distribution shift scenarios will be essential for confirming its utility. Additionally, we remain open to exploring alternative methods that could also effectively address these challenges.

\bibliography{main}

@inproceedings{
li2021tilted,
title={Tilted Empirical Risk Minimization},
author={Tian Li and Ahmad Beirami and Maziar Sanjabi and Virginia Smith},
booktitle={International Conference on Learning Representations, ICLR},
year={2021},
}

@inproceedings{
maity2023understanding,
title={Understanding new tasks through the lens of training data via exponential tilting},
author={Subha Maity and Mikhail Yurochkin and Moulinath Banerjee and Yuekai Sun},
booktitle={International Conference on Learning Representations, {ICLR} },
year={2023},
}

@article{JMLR:v8:sugiyama07a,
  author  = {Masashi Sugiyama and Matthias Krauledat and Klaus-Robert M{{\"u}}ller},
  title   = {Covariate Shift Adaptation by Importance Weighted Cross Validation},
  journal = {Journal of Machine Learning Research},
  year    = {2007},
  volume  = {8},
  number  = {35},
  pages   = {985--1005}
}

@article{SHIMODAIRA2000227,
title = {Improving predictive inference under covariate shift by weighting the log-likelihood function},
journal = {Journal of Statistical Planning and Inference},
volume = {90},
number = {2},
pages = {227-244},
year = {2000},
issn = {0378-3758},
author = {Hidetoshi Shimodaira},
}

@incollection{BIONDINI201529,
  title        = {An Introduction to Rare Event Simulation and Importance Sampling},
  author       = {Gino Biondini},
  booktitle    = {Big Data Analytics},
  editor       = {Venu Govindaraju and Vijay V. Raghavan and C.R. Rao},
  series       = {Handbook of Statistics},
  volume       = {33},
  pages        = {29--68},
  year         = {2015},
  publisher    = {Elsevier},
  issn         = {0169-7161},
}

@article{CHIRON2023109238,
title = {Failure probability estimation through high-dimensional elliptical distribution modeling with multiple importance sampling},
journal = {Reliability Engineering \& System Safety},
volume = {235},
pages = {109238},
year = {2023},
issn = {0951-8320},
author = {Marie Chiron and Christian Genest and Jérôme Morio and Sylvain Dubreuil},
}

@book{bandit,
  title     = {Bandit Algorithms},
  author    = {Lattimore, Tor and Szepesvári, Csaba},
  year      = {2020},
  publisher = {Cambridge University Press},
  address   = {Cambridge}
}

@book{zhang2023dive,
    title={Dive into Deep Learning},
    author={Zhang, Aston and Lipton, Zachary C. and Li, Mu and Smola, Alexander J.},
    publisher={Cambridge University Press},
    adress={Cambridge},
    year={2023}
}

@book{shift,
author = {Quionero-Candela, Joaquin and Sugiyama, Masashi and Schwaighofer, Anton and Lawrence, Neil D.},
title = {Dataset Shift in Machine Learning},
year = {2009},
isbn = {0262170051},
publisher = {The MIT Press}
}

@article{imbalance_bias,
author = {Wallace, Byron and Dahabreh, Issa},
year = {2013},
month = {10},
title = {Improving class probability estimates for imbalanced data},
volume = {41},
pages = {33–52},
journal = {Knowledge and Information Systems}
}

@INPROCEEDINGS{calibration,
  author={Pozzolo, Andrea Dal and Caelen, Olivier and Johnson, Reid A. and Bontempi, Gianluca},
  booktitle={2015 IEEE Symposium Series on Computational Intelligence}, 
  title={Calibrating Probability with Undersampling for Unbalanced Classification}, 
  year={2015},
  volume={},
  number={},
  pages={159-166},
  doi={10.1109/SSCI.2015.33}}

@ARTICLE{lin2017focal,
  author={Lin, Tsung-Yi and Goyal, Priya and Girshick, Ross and He, Kaiming and Dollár, Piotr},
  journal={IEEE Transactions on Pattern Analysis and Machine Intelligence}, 
  title={Focal Loss for Dense Object Detection}, 
  year={2020},
  volume={42},
  number={2},
  pages={318-327},
  doi={10.1109/TPAMI.2018.2858826}}

@inbook{conformal,
author = {Tibshirani, Ryan J. and Barber, Rina Foygel and Cand\`{e}s, Emmanuel J. and Ramdas, Aaditya},
title = {Conformal prediction under covariate shift},
year = {2019},
publisher = {Curran Associates Inc.},
address = {Red Hook, NY, USA},
booktitle = {Proceedings of the 33rd International Conference on Neural Information Processing Systems},
articleno = {227},
numpages = {11}
}

\end{document}